# MRI brain tumor segmentation using informative feature vectors and kernel dictionary learning

Seyedeh Mahya Mousavi, Mohammad Mostafavi*

*Department of Electrical Engineering, University of Zanjan, Zanjan, Iran*

**Abstract**:

This paper presents a method based on a kernel dictionary learning algorithm for segmenting brain tumor regions in magnetic resonance images (MRI). A set of first-order and second-order statistical feature vectors are extracted from patches of size $3 \times 3$ around pixels in the brain MRI scans. These feature vectors are utilized to train two kernel dictionaries separately for healthy and tumorous tissues. To enhance the efficiency of the dictionaries and reduce training time, a correlation-based sample selection technique is developed to identify the most informative and discriminative subset of feature vectors. This technique aims to improve the performance of the dictionaries by selecting a subset of feature vectors that provide valuable information for the segmentation task. Subsequently, a linear classifier is utilized to distinguish between healthy and unhealthy pixels based on the learned dictionaries. The results demonstrate that the proposed method outperforms other existing methods in terms of segmentation accuracy and significantly reduces both the time and memory required, resulting in a remarkably fast training process.

## 1. Introduction

In the field of medical imaging, segmentation plays a critical role in the identification and localization of pathologies, such as brain tumors. This process entails dividing an image into separate regions based on their distinct characteristics. Manual segmentation performed by human experts is prone to inter-observer variability and can be time-consuming. Therefore, there is a need for automated segmentation methods that can precisely and efficiently segment medical images without any human intervention [1]. Several methods have been proposed for brain MRI tumor segmentation, including thresholding, region growing [2], atlas-based methods [3], and machine learning-based approaches. Machine learning-based methods have gained popularity in recent years due to their ability to learn complex patterns in medical images and their potential to improve segmentation accuracy. D. Zikic et al. [4] presented a method for the automatic segmentation of brain tumor tissues using convolutional neural networks (CNNs). The proposed method involved preprocessing the input data, followed by training a CNN to classify each voxel of the input image into one of four tumor tissue classes. The limitations of the proposed method included the need for a large amount of training data, the computational complexity of training the CNN, and the challenge of handling class imbalance. Z. Jiang et al. [5] introduced a new technique that uses deep learning, specifically a two-stage cascaded U-Net, to segment brain tumors. The method involves segmenting the tumor in two stages, starting with a coarse segmentation followed by a finer segmentation, to improve the accuracy of the results. This approach was successful in accurately segmenting the different substructures of brain tumors. The proposed scheme by N. Bouchaour et al. in [6] utilized a large set of partial sub-images extracted from an MRI volume, which are then fed into an ensemble of convolutional neural networks (CNNs) to classify the central voxels into their respective tumor classes. By capturing local patterns around the central voxels, the use of partial sub-images enables faster training and prediction, facilitating efficient tumor diagnosis in real MRI-based scenarios. The experimental evaluations clearly showed that the proposed scheme achieves fast and accurate segmentation of brain tumors in pathological MRIs. In [7] a brain tumor segmentation method for multimodal MRI that combines deep semantic and edge information was presented. The proposed method comprises three modules: a semantic segmentation module utilizing the Swin Transformer for extracting semantic features with a shifted patch tokenization strategy, an edge detection module based on convolutional neural networks (CNNs) with an edge spatial attention block (ESAB) for feature enhancement, and a feature fusion module employing a multi-feature inference block (MFIB) based on graph convolution for effective fusion of semantic and edge features. The method demonstrated superior performance compared to several state-of-the-art brain tumor segmentation methods.

---

* Corresponding Author: Mohammad Mostafavi (mmostafavi@znu.ac.ir)

Furthermore, early research in the field of brain tumor segmentation also explored the use of dictionary learning-based methods, as described in [8]. The initial successes of these methods set the stage for the emergence of more advanced and refined techniques in recent years, including the proposed method in [9], which demonstrated a new method for automatic segmentation of brain MRI images using discriminative dictionary learning and sparse coding techniques. The proposed method achieves competitive results and is robust for different subject groups. The limitations of this method include the assumption that the training data is representative of all possible variations in the test data and its limited applicability to highly complex and heterogeneous tumors. In [10], a novel patch-based dictionary learning algorithm for automatic multi-label brain tumor segmentation was presented. The algorithm uses coupled dictionaries of grayscale brain tumor image patches and tumor labels. Although the proposed method demonstrated advantages in accuracy and robustness, it required a significant number of computational resources. S. Roy et al. [11] modified a patch-based tissue classification method from human brain MRI images that uses sparse dictionary learning and atlas priors. The main drawback of this method is its computational complexity. Yuhong et al. [12] presented an accurate brain tumor segmentation algorithm that combines sparse representation and Markov random field (MRF) to address the spatial and structural variability problems in brain tumor segmentation. The results show that the proposed method has some limitations, such as the need for manual selection of parameters and the sensitivity to noise in the images. In a remarkable work [13], a fully automatic brain tumor segmentation method based on kernel sparse coding, validated with 3D multiple-modality MRI, was described. Sparse coding is performed on feature vectors extracted from MRI image patches to code voxels, and morphological filtering is used to improve segmentation quality. Results showed that the proposed method has potential for differentiating healthy and pathological tissues. An automatic brain tumor segmentation method based on texture features and kernel sparse coding from FLAIR contrast-enhanced MRIs was presented in [14]. In this method sparse coding is performed on statistical eigenvectors extracted from MRIs, and kernel dictionary learning is used to construct adaptive dictionaries for healthy and pathological tissues. This method requires significant computational resources due to the large number of eigenvectors involved, making it a time-consuming process. An improved fuzzy C-means (FCM) algorithm with anti-noise capability for image segmentation, which combines dictionary learning with FCM clustering, proposed in [15]. The segmented images by the dictionary learning FCM (DLFCM) algorithm demonstrate better visual perception and clustering accuracy compared to other algorithms. The method stated in [16], described an automatic sparse constrained level set method to address the limitations of classic segmentation methods when segmenting brain tumors in MR images. The method constructs a sparse representation model of common characteristics of brain tumors' shapes and considers it a prior constraint in an energy function based on the level set method.

In this paper, we propose a method for brain MRI tumor segmentation using a kernel dictionary learning algorithm. First, the images are preprocessed to reduce noise and potential artifacts. Secondly, feature vectors, including first-order and second-order statistical features, are extracted from the preprocessed images. Then, a subset of discriminative feature vectors is selected using our proposed method based on the max-relevance and min-redundancy criteria with the Pearson correlation coefficient. In the next step, separate dictionaries are trained for tumor and normal tissue using the kernel k-SVD method. Finally, a linear classifier is employed to classify pixels as tumorous or normal based on the trained dictionaries.

## 2. The problem statements

### 2.1. Kernel Dictionary Learning:

Dictionary learning (DL) methods aim to design a set of basis elements, known as atoms, that can provide sparse representations for signals within a specific class. The algorithm for kernel DL iterates between updating the kernel dictionary and performing kernel sparse coding, which is used to update the dictionary in the next iteration. This process continues until the stopping criteria are met, which may be defined by a specific error threshold. In the kernel sparse coding stage, the Kernel Orthogonal Matching Pursuit (KOMP) algorithm is utilized, which employs a kernel function to map the data into a higher-dimensional space before calculating the sparse representation. This enables the algorithm to capture non-linear relationships between the data and the coefficients, potentially enhancing the accuracy of the sparse representation. The Kernel KSVD (KKSVD) algorithm is then employed to obtain the appropriate dictionary. In the subsequent figures, as described in [17], a concise description of each stage is provided.

**KOMP algorithm:**
**Input:** a signal $z$, a kernel function $\kappa$, a coefficient matrix $A$, and sparsity level $T_0$
**Output:** A coefficient vector $x \in \mathbb{R}^K$ with the most $T_0$ non-zero coefficients such that $\|\Phi(Y) - \Phi(Y)AX\|^2$ is minimized.
1. Initialize the variables: $s = 0$, $I_0 = \emptyset$, $x_0 = 0$, $v_0 = 0$
2. For each iteration $s = 1, 2, ..., T_0$, do the following:
   a) Compute the value of $\tau_i$ for each $i$ that is not in the index set $I_{s-1}$,
      $$\tau_i = \left(\mathcal{K}(z,Y) - v_s^T \mathcal{K}(Y,Y)\right) a_i, \quad \forall i \notin I_{s-1},$$
      where $[\mathcal{K}(i,j)] = [\kappa(y_i, y_j)]$
   b) Select the index $i_{max}$ that has the largest absolute value of $\tau_i$.
   c) Add $i_{max}$ to the index set $I_s$, which is the union of $I_{s-1}$ and $i_{max}$.
   d) Compute the coefficient vector $x_s$ using the subset of $A$ and $\kappa$ corresponding to the indices in $I_s$,
      where $x_s = \left(A_{I_s}^T \mathcal{K}(Y,Y) A_{I_s}\right)^{-1} \left(\mathcal{K}(z,Y) A_{I_s}\right)^T$.
   e) Compute the vector $v_s$ using the new value of $x_s$ and the corresponding subset of $A$, where $v_s = A_{I_s} x_s$
3. The output of the algorithm is a sparse vector $x$ that has $T_0$ non-zero coefficients at the indices in $I_s$ and zero coefficients everywhere else.

Figure 1. The kernel orthogonal matching pursuit algorithm.

> **KKSVD Algorithm:**
> **Input:** A set of signals $Y$, a kernel function $\kappa$
> **Task:** Find a dictionary via $A$ to represent these signals as sparse decompositions in the feature space.
> **Initialize**:
> - Set a random element of each column in $A^{(0)}$ to be $1$.
> - Normalize each column in the initial dictionary to a unit norm.
> - set iteration $J = 1.$
>
> Stage 1: sparce coding
> - Use the KOMP algorithm described in fig 1 to obtain the sparse coefficient matrix $X^{(J)}$ given a fixed dictionary $A^{(J-1)}$.
>
> Stage 2: Dictionary Update
> For each column $a_k^{(J-1)}$ in $A^{(J-1)}$, where $k = 1, \ldots, K,$ update it by
> - Define the group of examples that use this atom as:
>   $\omega_k = \{i | 1 \leq i \leq N, x_T^k(i) \neq 0\}$
> - Define the representation Error matrix by $E_k = \left(I - \sum_{j \neq k}^{K} a_j x_T^j\right)$
> - Restrict $E_k$ by choosing only the columns corresponding to $\omega_k$, and obtain $E_k^R$ as $E_k^R = E_k \Omega_k$.
> - Apply SVD decomposition to get $(E_k^R)^T \mathcal{K}(Y,Y)(E_k^R) = V \Delta V^T.$
> - Choose updated $a_k^{(J)} = \sigma_1^{-1} E_k^R v_1$, where $v_1$ is the first vector of $V$ corresponding to largest singular value $\sigma_1^2$ in $\Delta$.
>
> Set $J = J + 1$. Repeat from stage $1$ until a stopping criterion is met.
> **Output**: The final dictionary $A$ and the sparse coefficient matrix $X$.

**Figure 2. The kernel K-singular value decomposition algorithm.**

## 2.2. Brain tumor segmentation process:

The approach proposed in this paper for brain tumor segmentation can be categorized as a pixel classification method. The segmentation process involves several steps, starting with preprocessing the brain MRI dataset. Feature vectors are then extracted from the preprocessed images, followed by the selection of the most discriminative feature vectors. A kernel dictionary is trained using the selected feature vectors, and kernel sparse coding is performed. Finally, a classification stage is carried out to classify the pixels as tumors or non-tumors.

### 2.2.1. Preprocessing:

Preprocessing brain MRI images is crucial for ensuring reliable analysis. In this paper, we employ three key preprocessing steps. Firstly, we perform grayscale normalization, scaling the pixel values from 0 to 255. This normalization ensures that the intensities are within a standard range. Secondly, we apply a Gaussian filter to reduce Gaussian noise. Lastly, we utilize the active contour method to create an accurate mask. This mask effectively separates the relevant regions of interest from the background, enabling precise identification of valid partitioned areas. The mask assigns a value of 0 to the background and 1 to the valid data area, ensuring accurate delineation.

### 2.2.2. Feature extraction:

In this image analysis process, we apply a sliding 3×3 window to valid MRI image regions. For each pixel, we compute a feature vector based on the surrounding 3×3 region. This feature vector comprises both first-order statistical characteristics, such as the mean, variance, skewness, and kurtosis, as well as second-order statistical characteristics obtained using a gray-level co-occurrence matrix. The gray-level co-occurrence matrix is computed considering four different angles (0°, 45°, 90°, and 135°) to measure homogeneity, contrast, energy, and entropy. To construct a comprehensive feature vector, we combine the central pixel's gray value with the difference between the highest and lowest gray level values in the surrounding 3×3 region. This combination is performed alongside the calculated statistical features. As a result, a compact and informative 22-dimensional feature vector is formed. By encompassing both local pixel information and statistical characteristics, this representation effectively captures significant information about the region's properties, facilitating further analysis and processing of the image.

### 2.2.3. Sample selection:

In our brain MRI tumor segmentation project, we face a challenge due to the significant difference in size between the tumor and normal tissue regions in the images. This results in a substantial difference in the number of feature vectors extracted from each region, specifically, we had approximately 32,000,000 feature vectors for the normal tissue and around 6,000,000 feature vectors for the tumor region. This imbalance can negatively impact the training of dictionaries, potentially causing a bias towards the class with more feature vectors and leading to suboptimal performance in capturing the unique characteristics of the underrepresented class. Additionally, this issue results in a significant burden on processing resources and also requires a considerable amount of time and memory for processing. To overcome this issue, it is crucial to balance the number of feature vectors for both classes and focus on selecting the most discriminative ones. Therefore, we propose a sample selection method that leverages the correlation among samples to evaluate their representativeness and redundancy. The high correlation between a candidate sample and the remaining unselected feature vectors indicates a strong relationship, suggesting that the candidate sample can effectively represent the remaining feature vectors of the specific class. Conversely, a low correlation between the candidate sample and the currently selected feature vectors suggests low redundancy within the selected feature vector set. Additionally, if the candidate sample exhibits low correlation with feature vectors from other classes, it indicates a lack of close relationship with those feature vectors. Therefore, the selection criterion of the proposed method is based on the correlation analysis among feature vectors in their own specific class and within the. Let's consider the task of selecting two subsets of samples from two complete sets of feature vectors that respectively represent normal, $F_N = [f_1, f_2, ..., f_{32,000,000}] \epsilon \mathbb{R}^{L \times n_N}$, and tumor, $F_T = [f_1, f_2, ..., f_{6,000,000}] \epsilon \mathbb{R}^{L \times n_T}$, tissues. Where $L$ is the dimension of vector extracted in feature extraction step and $n_N$ and $n_T$ respectively are the number of feature vectors in each normal and tumorous class. Then the correlation matrix for both classes is denoted respectively as follows:

$$C_N = \begin{bmatrix} c_{1,1} & \cdots & c_{1,n_N} \\ \vdots & \ddots & \vdots \\ c_{n_N,1} & \cdots & c_{n_N,n_N} \end{bmatrix} \quad (1)$$

$$C_T = \begin{bmatrix} c_{1,1} & \cdots & c_{1,n_T} \\ \vdots & \ddots & \vdots \\ c_{n_T,1} & \cdots & c_{n_T,n_T} \end{bmatrix} \quad (2)$$

And the correlation matrix between both classes is defined as:

$$C_{N,T} = \begin{bmatrix} c_{1_N,1_T} & \cdots & c_{1_N,n_T} \\ \vdots & \ddots & \vdots \\ c_{n_N,1_T} & \cdots & c_{n_N,n_T} \end{bmatrix} \quad (3)$$

where;

$$C_{(i,j)} = \frac{cov\,(i,j)}{\sqrt{var(i)\,var(j)}} \quad (4)$$

$C_N$ and $C_T$ denote the correlation matrix of all the samples within their corresponding class and $C_{N,T}$ denotes the correlation matrix between two classes of feature vectors, and $C_{(i,j)}$ represents the feature vectors correlation between the i-th feature vectors $f_i$ and the j-th feature vectors $f_j$. It should be noted that we define that $C_{(i,j)}$ is proportional to the correlation, i.e., the higher the correlation between $f_i$ and $f_j$ is, the larger the value for $C_{i,j}$ is.

Assume that we aim to select $n$ samples out of the complete set of feature vectors of each class and have obtained $m$ samples ($m \in [0, n]$), to find the ($m + 1$)-th desired sample, we need to evaluate the representative ability and the degree of redundancy and separation of each unselected feature vectors. In the sample selection process, we aim to choose the most appropriate sample by scoring each vector in the unselected feature set. This scoring process consists of three stages.

Firstly, we measure the average correlation of a candidate sample with the other feature vectors in its corresponding unselected class. This allows us to assess how well the candidate sample aligns with the patterns and characteristics of its own class. Note that depending on whether we want to choose a sample from the normal or tumor class, $C_N$ or $C_T$ is used respectively. In the following equation $\Phi_U$ indicates the set of unselected feature vectors.

$$P_U^{(i)} = \frac{1}{n_T - m - 1} \sum_{j \in \Phi_U} c_{i,j} \quad (5)$$

Secondly, we evaluate the average correlation of the candidate sample with the other selected samples. This step helps us minimize redundancy within the selected samples by considering how well the candidate sample complements the existing set. In the following equation $\Phi_S$ indicates the set of selected samples.

$$P_S^{(i)} = \frac{1}{m} \sum_{j \epsilon \Phi_S} c_{i,j} \tag{6}$$

Lastly, we examine the average correlation of the candidate sample with the feature vectors of the other class. This enables us to determine the discriminatory power of the candidate sample in distinguishing between different classes. $C_{N,T}$, which defines correlation between two class, is used. Note that depending on whether we want to choose a sample from the normal or tumor class, the average differs. If the candidate sample belongs to normal class $r = n_T$ and if the candidate sample belongs to tumor class $r = n_N$ and the average is defined as:

$$P_O^{(i)} = \frac{1}{r} \sum_{j=1}^{r} c_{i,j} \tag{7}$$

and if the candidate sample belongs to tumor class, $r = n_N$ and the average is defined as:

$$P_O^{(i)} = \frac{1}{r} \sum_{i=1}^{r} c_{i,j} \tag{8}$$

Lastly, the following equation calculates the total score of each candidate sample.

$$P_T = P_U^{(i)} - P_S^{(i)} - P_O^{(i)} \tag{9}$$

Ultimately, the candidate feature vector with the maximum score is selected in each iteration, ensuring that the chosen samples contribute positively to the classification task, whether it is for the normal or tumor class. By repeating this process until a predetermined number of selected feature vectors is reached, we can iteratively identify the most suitable feature vectors.

**2.2.4. Segmentation process:**

step1: Utilize a radial basis function (RBF) as the kernel function, denoted as $\kappa(x, y) = \exp(-\gamma \|x - y\|^2)$, where $\gamma$ is a parameter value set to 0.35. The RBF kernel allows us to transform the original data representation into a higher-dimensional non-linear space.

step2: Train two over-complete kernel dictionaries, namely $D_N$ and $D_T$, using the KOMP and KKSVD algorithms (Fig1. and Fig2.), respectively. Each dictionary should be train with an equal number of the most discriminative selected feature vectors, using proposed sample selection method, for the normal tissue and the tumorous region. These dictionaries are designed to capture the specific characteristics and patterns present in each area, allowing for more accurate representation and analysis of the corresponding tissue types.

step3: To determine the class of a given test feature vector $f_i'$, we employ KKSVD algorithm and the dictionaries $D_N$ and $D_T$ separately to make it sparse and obtain $f_i^N$ and $f_i^T$ respectively. These sparse representations are then used as input for a classification model based on root mean square error (RMSE) used in order to classify the category of the given $f_i$ as follows:

$$¥(f_i) = \begin{cases} Normal, & if\ \text{RMSE}\ (f_i', f_i^N) > \text{RMSE}\ (f_i', f_i^T) \\ tumorous, & elsewhere \end{cases} \quad (10)$$

where $\text{RMSE}\ (f_i', f_i) = \sqrt{MSE(f_i', f_i)} = \sqrt{E((f_i', f_i)^2)}$

## 3. Main results

### 3.1. Data origin:

The data resource for this research is BRATS 2020 [18-20] dataset. The dataset comprises 120 records of glioma patients, including both high-grade and low-grade cases. It includes four modalities of MRI scans: T1, T2, T1C, and FLAIR. Manual segmentation of the tumors, performed by experienced physicians and experts, is available as corresponding label data. For this study, a subset of 36 records was selected from the dataset, where the patients' age was younger than 70 and their survival days were less than 365. These records were used as training data. Additionally, another subset of 11 records with similar conditions was used as validation data. The purpose of the validation data was to evaluate the performance and generalization ability of the developed model. The evaluation method focused on FLAIR-weighted MRI images to identify the entire tumor region and compared the results with the provided standard segmentation data. The segmentation process has been implemented on three axes of brain images, namely X, Y, and Z. For each axis, two dictionaries were trained, and the segmentation process was performed separately on each axis.

## 3.2. The optimization of parameter values:

Parameter selection plays a pivotal role in dictionary learning algorithms, with the number of atoms in the kernel dictionary (K) and the number of selected feature vectors (n) in the sample selection step serving as prime examples. Among the ten randomly chosen numbers of atoms, namely K = [11, 32, 98, 133, 189, 242, 287, 354, 426, 493], as depicted in Figure.1, we observed notable variations in segmentation accuracy. Particularly, our proposed method yielded seven results, where the dictionary with 32 atoms exhibited the highest accuracy, surpassing the other tested numbers. In the selection of the optimal number of feature vectors during the sample selection stage, we observed that the number of extracted feature vectors differed between tumorous and normal tissue due to the disparity in tumor volume compared to healthy tissue. To mitigate potential bias, we aimed to choose an equal number of samples. Specifically, we considered sample counts of 5,000,000, 2,500,000, 1,250,000, 625,000, 312,500, 156,250, and 78,125, and the performance was evaluated at each step. The effects of reducing the number of samples on the mean of four calculated evaluation metrics are shown in Table 1. As the number of samples was reduced from 5,000,000 to 156,250, the evaluation metrics consistently improved, indicating the effectiveness of the proposed sample selection method in utilizing discriminative feature vectors and eliminating redundancy. However, a further decrease in the number of samples to 78,125 resulted in a significant deterioration of the results, likely due to the loss of crucial discriminatory information.

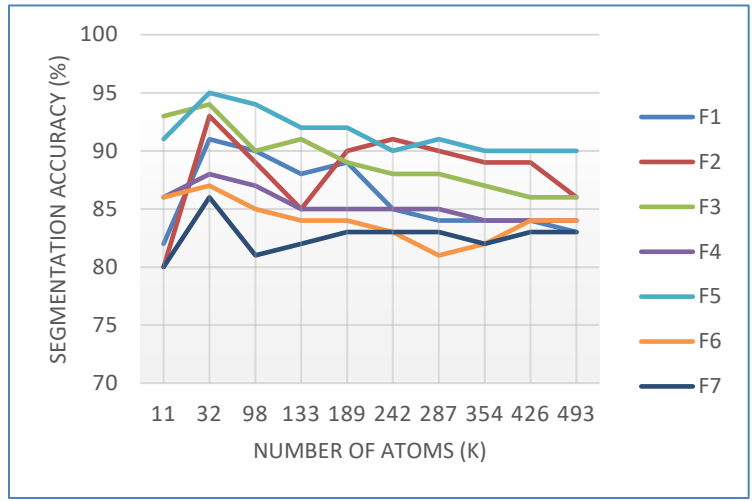

**Figure 3.** The effect of choosing different number of atoms in segmentation accuracy for seven patients.

**Table 1.** The effect of reducing number of feature vectors in segmentation result.

| Number of features  Metric | 5,000,000 | 2,500,000 | 1,250,000 | 625,000 | 312,500 | 156,250 | 78,125 |
|---|---|---|---|---|---|---|---|
| Dice | 0.8987 | 0.8992 | 0.9021 | 0.9033 | 0.9040 | 0.9048 | 0.6038 |
| Jaccard | 0.8043 | 0.8092 | 0.8106 | 0.8151 | 0.8239 | 0.8321 | 0.5830 |
| Sensitivity | 0.9311 | 0.9354 | 0.9479 | 0.9400 | 0.9438 | 0.9455 | 0.6203 |
| Specificity | 0.9990 | 0.9991 | 0.9992 | 0.9993 | 0.9993 | 0.9994 | 0.5865 |

### 3.3. Segmentation results:

The segmentation results achieved in this study provide compelling evidence of the success of our proposed method for brain tumor segmentation in MRI images. The Dice score, utilized as a precision indicator for segmentation, exhibits a high level of accuracy in segmenting brain tumor regions, with values ranging from 0.8843 to 0.9367, as shown in Table2. Additionally, the other three indexes also demonstrate a remarkable level of accuracy in delineating tumor regions. These findings emphasize the effectiveness of our approach in accurately segmenting brain tumors and validate the success of our method. Furthermore, the proposed sample selection method significantly improves the results obtained from the segmentation process, both in terms of computational load (reduced memory usage) and the time required for training dictionary calculations. This improvement is demonstrated in Table 3, where the results reveal that selecting an optimal number of samples at 156,250 ensures robust and reliable segmentation outcomes. These findings highlight the effectiveness of our sample selection approach in enhancing computational efficiency and achieving accurate segmentation results. The visual segmentation results obtained by training dictionaries using 156,250 samples with the KOMP and KKSVD algorithms are presented in Fig4. The figure displays the original MRI image, the available label indicating the tumor region, and the segmented area generated by our proposed method in three axis of brain image. These visual results provide a clear representation of the segmentation outcomes achieved through our approach, showcasing the effectiveness and accuracy of our method in delineating tumor regions in MRI images.

**Table 2. Performance of proposed method in four indexes.**

| | 156,250 selected samples | | | |
|---|---|---|---|---|
| case number | Jaccard | Dice | Sensitivity | Specificity |
| P1 | 0.8061 | 0.8891 | 0.9234 | 0.9992 |
| P2 | 0.8438 | 0.9078 | 0.9567 | 0.9996 |
| P3 | 0.8713 | 0.9231 | 0.9678 | 0.9997 |
| P4 | 0.7627 | 0.8765 | 0.9005 | 0.9990 |
| P5 | 0.8314 | 0.9019 | 0.9445 | 0.9995 |
| P6 | 0.8776 | 0.9254 | 0.9789 | 0.9998 |
| P7 | 0.8089 | 0.8923 | 0.9321 | 0.9993 |
| P8 | 0.7945 | 0.8843 | 0.9123 | 0.9991 |
| P9 | 0.8845 | 0.9367 | 0.9902 | 0.9999 |
| P10 | 0.8539 | 0.9176 | 0.9598 | 0.9997 |
| P11 | 0.8192 | 0.8982 | 0.9345 | 0.9994 |
| Mean | 0.8322 | 0.9048 | 0.9455 | 0.9995 |
| Std. | 0.0382 | 0.0190 | 0.0280 | 0.0003 |

Table 3. Required time for training kernel dictionary with different numbers of samples.

| Number of samples | 5,000,000 | 2,500,000 | 1,250,000 | 625,000 | 312,500 | 156,250 |
|---|---|---|---|---|---|---|
| Time required for training kernel dictionary | 7146 min | 5087min | 2382min | 1053min | 595min | 423min |

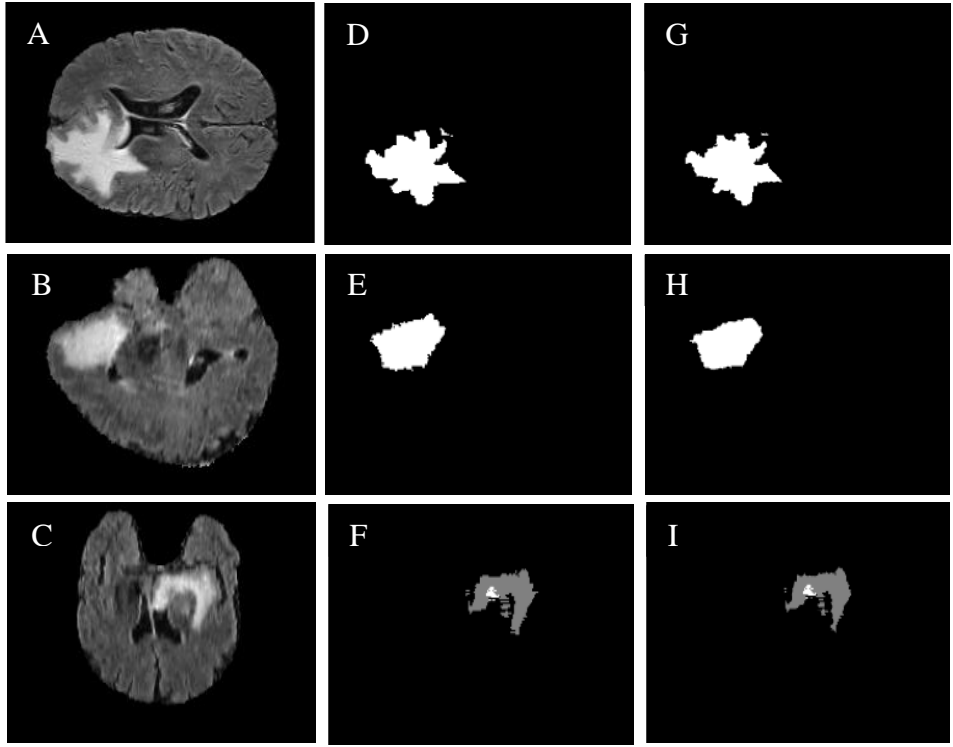

**Figure 4. Segmentation Results: Original MRI (A, B, C) | Expert Annotations (D, E, F) | Proposed Algorithm (G, H, I)**

## 4. Discussion and conclusion:

The proposed brain tumor segmentation method utilizes kernel dictionary learning and selected texture-based extracted features, incorporating the Pearson correlation coefficient. Preprocessing steps involve applying a Gaussian filter and grayscale level normalization to images from the BRATS2020 dataset. Spatial features with first and second-order statistical features are then extracted, resulting in a 22-dimensional feature vector. To address class imbalance, a sample selection method based on the Pearson correlation coefficient is introduced, which assigns scores to candidate samples for selection based on correlations within and between classes and redundancy. Two separate kernel dictionaries are trained using the KOMP and KKSVD algorithms for tumor and healthy classes, respectively. A linear RMSE-based classifier is employed for pixel-by-pixel segmentation of test images, effectively separating the tumor region. The proposed approach enhances the accuracy of tumor region boundaries while reducing computation time and memory requirements through the discriminative sample selection stage.

Compared with other segmentation methods presented in references [14, 18] the method proposed in this study demonstrates superior performance in terms of both dice score and sensitivity, as highlighted in Table 4. The dice score is a widely-used measure of overlap between the segmented region and the ground truth, indicating the accuracy of the segmentation. A higher dice score signifies a better alignment between the predicted and actual tumor regions. Similarly, sensitivity measures the ability of the segmentation method to correctly identify true positive tumor pixels. By achieving the highest dice score and sensitivity values, our proposed method outperforms the referenced methods in accurately capturing the tumor region and detecting tumor pixels, which is crucial for effective brain tumor segmentation. These results demonstrate the efficacy and potential of our approach in achieving more accurate and reliable segmentation outcomes. It should be noted that the specificity metric value for all the mentioned methods in Table 4 is equal to 1.00, which is intentionally omitted from the table to prioritize the highlighting of the crucial distinguishing values among these methods. Regarding the segmentation performance, it is observed that images with minimal noise achieve a maximum accuracy of 96%. The proposed method utilizes a Gaussian filter during the preprocessing stage to effectively reduce image noise and rectify inconsistencies in MRI gradation, resulting in improved segmentation accuracy.

In future work, we can investigate additional techniques to mitigate class imbalance, explore feature selection and dimensionality reduction methods, evaluate alternative dictionary learning algorithms, test the method on different datasets and medical imaging modalities, and conduct clinical validation studies. These future directions can enhance the method's performance, robustness, and applicability in real-world scenarios, ultimately contributing to the advancement of brain tumor segmentation in medical imaging.

**Table 4. Comparative Analysis of Segmentation Methods: Effectiveness and Performance Metrics**

|  | Dice | sensitivity |
|---|---|---|
| Tong | 0.88 | 0.90 |
| Hamamci | 0.82 | 0.86 |
| Zikic | 0.91 | 0.88 |
| Bauer | 0.87 | 0.84 |
| The proposed method | 0.92 | 0.90 |

**Conflict of interest statement:** The authors declare no conflict of interest in preparing this article.

**Funding Acknowledgement:** This research received no specific grant from any funding agency in the public, commercial, or not-for-profit sectors.